# A Survey on Two Dimensional Cellular Automata and Its Application in Image Processing

Deepak Ranjan Nayak
Dept. of CSE, College of Engineering and Technology, Bhubaneswar, India

depakranjannayak@gmail.com

Prashanta Kumar Patra
Dept. of CSE, College of Engineering and Technology, Bhubaneswar, India

pkpatra@cet.edu.in

Amitav Mahapatra
Dept. of CSE, College of Engineering and Technology, Bhubaneswar, India

amitavmahapatracse@cet.edu.in

## ABSTRACT

Parallel algorithms for solving any image processing task is a highly demanded approach in the modern world. Cellular Automata (CA) are the most common and simple models of parallel computation. So, CA has been successfully used in the domain of image processing for the last couple of years. This paper provides a survey of available literatures of some methodologies employed by different researchers to utilize the cellular automata for solving some important problems of image processing. The survey includes some important image processing tasks such as rotation, zooming, translation, segmentation, edge detection, compression and noise reduction of images. Finally, the experimental results of some methodologies are presented.

## 1. INTRODUCTION

The concept of CA was initiated in the early 1950's by J. Von Neumann and Stan Ulam [1, 2]. Afterwards, Stephen Wolfram developed the CA theory [3]. Cellular automata (CA) are now becoming an attractive area for researchers of various fields due to its parallel nature. It is not only used in the field of engineering but also used in every field of sciences. The reason behind the popularity of cellular automata can be traced to their simplicity, and to the enormous potential they hold in modeling complex systems, in spite of their simplicity [4].

Digital image processing in many applications is considered a real-time process in which the process speed is very important. Therefore, the parallel algorithms in image processing are much more important compared with serial algorithms [28]. CA are widely used by researchers in the domain of image processing. So, CA can be used as a parallel method for any image processing task [12].

A cellular automaton (CA) is a collection of cells arranged in an N-Dimensional (N-D) lattice, such that each cell's state changes as a function of time according to defined set of rules that includes the states of the neighboring cells. Typically, the rule for updating the cells state is the same for each cell, it does not change over time and it is applied to the whole grid simultaneously. That is, the new state of each cell, at the next time step, depends only on the current state of the cell and the states of the cells in its neighborhood. All cells on the lattice are updated synchronously. Thus, the state of the entire lattice advances in discrete time steps. It is clear that the concept of parallelism is implicit to cellular automata [12, 26, 27].

Two most common types of CA used by different authors are: one dimensional CA (1D CA) and two dimensional CA (2D CA). If the grid is a linear array of cells, is called 1D CA and if it is a rectangular or hexagonal grid of cells then it is called 2D CA. A CA with one central cell and four near neighborhood cells is called a von Neumann/Five neighborhood CA whereas a CA having one central cell and eight near neighborhood cells is called Moore/Nine neighborhood CA [27].

An image can be viewed as a two dimensional CA where each cell represents a pixel in the image and the intensity of the pixel is represented by the state of that cell. The states of the cells are updated synchronously at a discrete time step. So the time complexity to do any image processing task is the least [5, 12]. Due to this kind of behavior of CA model influences a large application in image processing such as image restoration, enhancement, segmentation, compression, feature extraction and pattern recognition.

Many standard algorithms for most of the image processing tasks have already been developed by different researchers in the last few decades. But some researchers later have used CA to solve the same problem and found to be better than conventional methods.

The survey paper has been laid out as follows. The next section introduces the concept of cellular automata. Section 3 highlights a survey of different types of cellular automata structures proposed over twenty five years. A wide variety of applications of cellular automata in image processing encountered by different researchers are presented in Section 4. Section 5 shows experimental results of some applications. Finally, the paper is concluded in the last section.

## 2. CONCEPT OF CELLULAR AUTOMATA

Cellular automata consists of regular grid of cells in which each cell can have finite number of possible states. The state of a cell at a given time step is updated in parallel and determined by the previous states of surrounding neighborhood of cells with the help of a specified transition rule (as shown in Figure 1). Thus, the rules of the CA are local and uniform.

In effect, each cell as shown in Figure 2, consists of a storage element (D flip-flop) and a combinational logic (CL) implementing the next-state function. The combinational logic is called the rule of the CA. The algorithm used to compute the next cell state is referred to as the CA local rule. The state of each cell is updated simultaneously at discrete time steps based on the states in its neighborhood at the preceding time step [8].

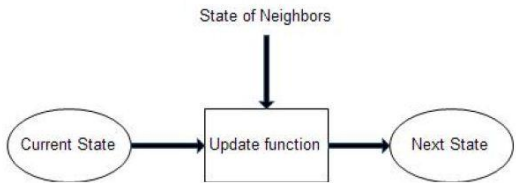

**Fig 1: State transition depend on neighborhood state [8]**

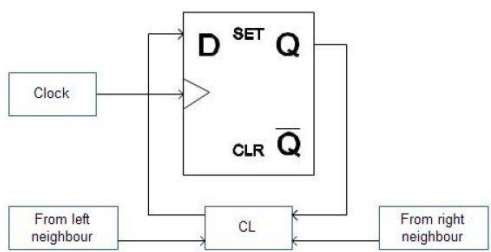

**Fig 2: A Typical CA cell [8]**

For a specific problem based on CA, we have to define some important terms: the lattice geometry, neighborhood size, boundary conditions, initial conditions, state set and transition rules [12]. These terms are all described in the following. There are one dimensional, two dimensional and three dimensional CA used to solve different type of problems. One dimensional CA (1D CA) consists of linear arrays of cells whereas in two dimensional CA (2D CA), cells are arranged in a rectangular or hexagonal grid with connections among the neighboring cells, which is depicted in Figure 3 and 4 respectively.

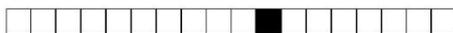

**Fig 3: Structure of 1D CA**

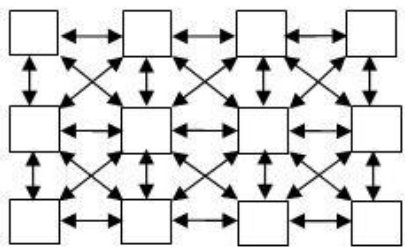

**Fig 4: Structure of 2D CA**

A CA can be represented with five- tuple, $C = \{L, N, Q, \delta, q0\}$; where $L$ is the regular lattice of cells, $Q$ is the finite set of states, $q0$ is called the initial state and $q0 \in Q$, $N$ is a finite set (of size $n = |N|$) of neighborhood indices such that for all $r \in L$, for all $c \in N: r + c \in L$ and $\delta: Q^n \to Q$ is the transition function [26].

For a 3- neighborhood 1D CA, the transition function can be represented as

$$qi(t + 1) = \delta\ (qi(t), qi - 1(t), qi + 1(t))$$

where $qi(t + 1)$ and $qi(t)$ denotes the state of the $i^{th}$ cell at time $t + 1$ and $t$ respectively, $qi - 1(t)$ and $qi + 1(t)$ represents the state of the left and right neighbor of the $i^{th}$ cell at time $t$, and $\delta$ is the next state function or the transition rule.

Since the digital image is a two-dimensional array of pixels, so we are mainly focused on two- dimensional CA model.

## 2.1 Neighborhood Structure

The neighborhood of a cell, called the core cell (or central cell), made up of the core cell and those surrounding cells whose states determine the next state of the core cell. There are different neighborhood structures for cellular automata. The two most commonly used neighborhoods are Von Neumann and Moore neighborhood, shown in Figure 5.

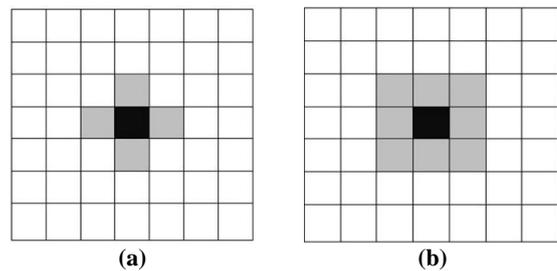

**(a)**        **(b)**

**Fig 5: Neighborhood model (a) Von Neumann, (b) Moore**

Von Neumann neighborhood has five cells, consisting of the cell and its four immediate non-diagonal neighbors and has a radius of 1. The radius of a neighborhood is defined to be the maximum distance from the core cell, horizontally or vertically, to cells in the neighborhood.

Moore neighborhood has nine cells, consisting of the cell and its eight surrounding neighbors and has a radius of 1. Extended Moore neighborhood composed of the same cells as the Moore neighborhood, but the radius of neighbourhood is increased to 2 [26].

## 2.2 Boundary Conditions

Most popular boundary conditions are null boundary and periodic boundary conditions which are used when a transition rule is applied to the boundary cells of CA. In null boundary conditions, the extreme cells are connected to logic 0- state and in periodic boundary conditions the extreme cells are adjacent to each other.

Some other types of boundaries are fixed, adiabatic and reflexive which are used to solve some specific type problems. If extreme cells are connected to any fixed state value, then it is termed as fixed boundary condition. In adiabatic boundary condition extreme cells are connected to its replicate state and in reflexive boundary condition mirror states replace the extreme cells [27].

## 2.3 Rule Description of 1D CA

Elementary cellular automata have two possible values for each cell (0 or 1), and rules that depend only on nearest neighbor values. As a result, the evolution of an elementary cellular automaton can completely be described by a table specifying the state a given cell will have in the next generation based on the value of the cell to its left, the value the cell itself, and the value of the cell to its right [4].

Since there are $2 * 2 * 2 = 2^3 = 8$ possible binary states for the three cells neighboring a given cell, there are a total of $2^8 = 256$ 1D cellular automata rules, each of which can be indexed with an 8-bit binary number. These are generally known as Wolfram's 256 1D CA rules (i.e. from Rule 0 to Rule 255). Two of the rules are presented in Table 1.

General formulae for computing number of uniform CA rules:

$$Number\ of\ Uniform\ CA\ rules = State^{State^{Neighborhood}}$$

**Table 1. Rule 90 and Rule 150**

| Neighborhood State | 111 | 110 | 101 | 100 | 011 | 010 | 001 | 000 | |
|---|---|---|---|---|---|---|---|---|---|
| Next State | 0 | 1 | 0 | 1 | 1 | 0 | 1 | 0 | Rule 90 |
| Next State | 1 | 0 | 1 | 0 | 1 | 0 | 1 | 1 | Rule 171 |

If the next state function of a state is expressed in the form of a truth table, then the decimal equivalent of the output is called rule number of the CA.

The top row gives all the eight possible states of the 3 neighboring cells at the time constant t, while the second and third row give the corresponding states of the $i^{th}$ cell at time instant t+1 for two illustrative CA rules. The second row taken as a binary number and converted into its decimal representation, is the rule number 90. Similarly, the third row corresponds to the rule number 171. We can represent all the 256 rules in the above manner.

### 2.4 Rule Description of 2D CA
In 2D Nine Neighborhood CA the next state of a particular cell is affected by the current state of itself and eight cells in its nearest neighborhood. Such dependencies are accounted by various rules. A specific rule convention that is adopted in [9] is represented in Table 2.

**Table 2. Two dimensional CA rule convention**

| | | |
|---|---|---|
| 64 | 128 | 256 |
| 32 | 1 | 2 |
| 16 | 8 | 4 |

The central box represents the current cell (i.e. the cell being considered) and all other boxes represent the eight nearest neighbors of that cell. The number within each box indicates the rule number characterizing the dependency of the current cell on that particular neighbor only. Rule 1 characterizes dependency of the central cell on itself alone whereas such dependency only on its top neighbor is characterized by rule 128, and so on.

For two state nine neighborhood CA, there are $2^{2^9}$ possible rules exist. Out of them only $2^9 = 512$ are linear rules that is, the rules which can be realized by EX-OR operations only and the rest of the $2^{2^9}$- $2^9$ rules are non-linear which can be realized by all possible operations of CA. Taking XOR operation (/s) among nine basic rules, we get other 502 linear rules (excluding Rule₀).

The example shown in the following illustrates how a composite Rule₄₄₉ is calculated using basic rules and XOR operations.

Example: Rule₄₄₉ is expressed in terms of basic rule matrices as follows:

$$Rule449 = Rule256 \oplus Rule128 \oplus Rule64 \oplus Rule1$$

### 2.5 Relationship of 2D CA with image
An image may be described as a two-dimensional function I.
$$I = f(x, y)$$
Where $x$ and $y$ are spatial coordinates. Amplitude of $f$ at any pair of coordinates $(x, y)$ is called intensity $I$ or gray value of the image. When spatial coordinates and amplitude values are all finite, discrete quantities, the image is called digital image. The digital image $I$ is represented by a single 2- dimensional integer array for a gray scale image and a series of three 2-dimensional arrays for each color bands [5]. As the digital image is a two-dimensional array of m×n pixels, so we are interested in two- dimensional CA model. An image is viewed as a two dimensional CA where each cell represents a pixel in the image and the intensity of the pixel is represented by the state of that cell [12, 27]. The color values of the pixels are updated synchronously at a discrete time step. So very less time is required to solve any image processing task.

## 3. TYPES OF CA
This section highlights different variations of CA which have been proposed by different researchers to ease the design and modelling of complex systems.

### 3.1 Uniform CA
The Uniform Cellular Automata have been presented by Nandi et al. [33]. If the same rule is applied to all the cells in a CA, then the CA is said to be uniform or regular CA.

### 3.2 Hybrid CA
Hybrid CA have been explored by Anghelescu et al. [34]. If in a CA the different rules are applied to different cells, then the CA is said to be hybrid CA.

### 3.3 Null Boundary CA
In [35], the null boundary Cellular Automata is proposed. A CA said to be a null boundary CA if both the left and right neighbor of the leftmost and rightmost terminal cell is connected to logic 0.

### 3.4 Periodic Boundary CA
The periodic boundary Cellular Automata have been explored by Anghelescu et al. [34]. In periodic boundary CA the rightmost cell as the left neighbor of leftmost cell and similarly, the leftmost cell is considered as the right neighbor of rightmost cell forming a structure like a circular linked list.

### 3.5 Linear CA
The linear Cellular Automata have been proposed by Nandi et al. [33]. If the rule of CA involves only XOR logic then it is called the linear rules. A CA with all the cells having linear rules is called linear CA. Most of the image processing tasks are done using linear CA.

### 3.6 Non-linear CA
Das et al. [36] introduces the concept of non-linear CA. Non-linear CA uses all possible logic available in CA.

### 3.7 Complement CA
Nandi et al. proposed a new type of CA called complement Cellular Automata [33]. If the rule of CA involves only XNOR logic then it is called the complement rules. A CA with all the cells having complements rules is called complement CA.

### 3.8 Additive CA
The additive Cellular Automata have been introduced by Nandi et al. [33]. A CA having a combination of XOR and XNOR rules is called Additive CA.

### 3.9 Programmable CA

The programmable Cellular Automata have been developed by Anghelescu et al. [34]. A CA is called programmable CA [7] if it employs some control signals. By specifying values of control signal at run time, programmable CA can implement various function dynamically.

### 3.10 Reversible CA

Seredynski et al. introduced the reversible Cellular Automata in [37]. A CA is said to be reversible CA in the sense that the CA will always return to its initial state. The interesting property of being the reversible which means not only forward but also reverse iteration is possible. Using reversible rule it is always possible to return to an initial state of CA at any point.

### 3.11 Generalized Multiple Attractor CA

The special class of CA, called as GMACA [38], is employed for the design of desired CA models, evolved through an efficient implementation of genetic algorithm, are found to be at the edge of chaos. Cellular automata are mathematical idealizations of complex systems in discrete space and time.

### 3.12 Fuzzy CA

The fuzzy Cellular Automata have been proposed by Maji et al. [22]. Fuzzy CA means CA employed with fuzzy logic. A special class of CA referred to as Fuzzy CA (FCA) [39] is employed to design the pattern classifier. In FCA all states of a cell and local transition function (rules) are fuzzy.

## 4. CA IN IMAGE PROCESSING

Two dimensional CA algorithms are widely used in image processing as its structure is similar to an image. This section presents a survey of previously published works employed by different researchers in the field of cellular automata with application in image processing. The survey is categorized into the following sub-headings.

### 4.1 Translation

Translation of images means moving the image in all the directions. Generally, we are able to translate images in x (left, right) and y (up, down) directions. Translation of images using CA includes diagonal movement of images. Choudhury et al. have applied eight basic two dimensional CA rules namely $Rule_2$, $Rule_4$, $Rule_8$, $Rule_{16}$, $Rule_{32}$, $Rule_{64}$, $Rule_{128}$, $Rule_{256}$ to any images for some iterations and found that these rules causes translation of images in all directions [9]. Their works are summarized in Table 3.

**Table 3. Translation of images using basic 2D CA rules [9]**

| Rules | Direction of Translation of Images |
|---|---|
| $Rule_2$ | Left |
| $Rule_{32}$ | Right |
| $Rule_8$ | Top |
| $Rule_{128}$ | Bottom |
| $Rule_4$ | Top- Left (Diagonal) |
| $Rule_{16}$ | Top- Right (Diagonal) |
| $Rule_{64}$ | Bottom- Right (Diagonal) |
| $Rule_{256}$ | Bottom- Left (Diagonal) |

An extension of the previous work was done by Qadir et al. [47]. They used twenty five neighborhood concept instead of nine neighborhood for translation of images. That is they have applied twenty four fundamental rules to any images and got satisfactory results. The main idea of their work was to move the images in more directions in addition to the previous one so that it can be used in gaming applications. The survey on translation using CA concludes that the number of iteration required for twenty five neighborhood model are less than nine neighborhood model.

### 4.2 Rotation

Rotation of images through an arbitrary angle is only possible by using hybrid CA [40]. He has illustrated that how extremely simple 2D CA rules can be used to rotate the images through an angle. Two rules namely $Rule_{136}$ and $Rule_{34}$ have been applied to an image under null boundary conditions to rotate the image by an angle $\pi$ about x axis and y axis respectively.

### 4.3 Zooming

In zooming there are two operations, zooming in and zooming out. Chaudhry et al. employed hybrid CA concept for zooming of symmetric images [9]. For zooming in they have applied $Rule_2$, $Rule_{32}$, $Rule_8$ and $Rule_{128}$ in four different regions of images respectively and for zooming out $Rule_{32}$, $Rule_2$, $Rule_{128}$, $Rule_8$ are applied respectively.

### 4.4 Thinning

Thinning is a special case of scaling and is an important procedure in image analysis. Thinning horizontal and vertical blocks using 2D hybrid CA has been illustrated by Chaudhry et al. [9]. They applied $Rule_{32}$, $Rule_2$, $Rule_1$ and $Rule_1$ in four different regions of images respectively for horizontal thinning and, $Rule_1$, $Rule_1$, $Rule_{128}$, $Rule_8$ for vertical thinning.

### 4.5 Edge Detection

Edge detection is one of the most fundamental approach in image processing. In an image, we generally concentrate on objects rather than on the background of an image. Hence objects are important. Edges characterize boundaries of objects. These are significant local changes of intensity in an image [5, 27]. Based on the above idea many algorithms have been developed for edge detection, however this is still a challenging and an unsolved problem in image processing.

Edge detection based on gradient operators and Laplacian operators requires much computing time. With an increasing demand for high speed real time image processing the need for parallel algorithms instead of serial algorithms is becoming more important. As an intrinsic parallel computational model, cellular automata (CA) can fulfill this need [26]. Different CA models were used by authors over the last few years for performing edge detection. These are summarized as follows.

S. Wongthanavasu et al. (2003), proposed a simple CA rule for edge detection [10], and an asynchronous CA model is presented by Scarioni in 1998 for the same task [11]. In 2004 Chang et al. introduced a new method of edge detection of gray images using CA [12]. They have considered nine neighborhood structures with periodic boundary condition. An orientation information measure is used to deal with the original grayscale matrix of the image. P L Rosin proposed a different approach on training binary CA for image processing task in the year 2006 [13]. Rather than use an evolutionary approach such as genetic algorithms, a deterministic method was employed, namely sequential floating forward search (SFFS). But the work was only dealt with processing of binary images. Later on, he extends the work to deal with gray images effectively [14]. A new approach for edge recognition based on the combinations of CA and a traditional method of image processing is proposed by Chen and Hao, where they used the concept of boundary operator to represent the state of a cell, and the local rule is defined based on prior knowledge [15]. Lee et al. (2010) proposed the concept of using cellular automata and adapted two algorithms for edge detection in hyper spectral images. The authors developed two CAs to analyze the image: an edge detection CA and a post-processing CA (that implements morphological operations for denoising the edges). Results demonstrated the CA method to be very promising for both unsupervised and supervised edge detection in hyperspectral imagery [16]. In 2011, a new method of segmentation has been proposed by Safia et al. which is an edge detector based on continuous cellular automaton for both binary images and grayscale images. The method presented the advantage of facilitating the evolution of cellular automata until the final configuration, without the tedious process of searching manually the rules set which leads to the desired result [25].

In the last couple of years, some researchers applied evolutionary algorithms (such as GA, PSO) to CA for generating a best rule to perform the edge detection task. Kazar et al. in 2011 used genetic algorithms with CA for image segmentation and noise filtering [17]. In 2012, a meta-heuristic PSO was introduced by S. Djemame to find out the optimal and appropriate transition rules set of CA for edge detection task. The efficiency of the method was very promising [18]. The concept of fuzzy logic is also somewhat combined with CA for the edge detection task. In 2004, Wang Hong et al. proposed a novel image segmentation arithmetic using fuzzy cellular automata (FCA) [19]. A new improved edge detection algorithm of fuzzy CA is introduced by Ke Zhang in 2007. It has been proved that, the method has great detections effect [20]. More and Patel recently used fuzzy logic based image processing for accurate and noise free edge detection and Cellular Learning Automata(CLA) for enhance the previously detected edges with the help of the repeatable and neighborhood considering nature of CLA [21].

Nayak et al. [26] presented a novel method for edge detection of binary images based on two dimensional twenty five neighborhood cellular automata. The method considers only some pattern of linear rules of CA for extraction of edges under null boundary condition. The performance of this approach was compared with some existing edge detection techniques. The experimental results published in the paper illustrates that the proposed method to be very promising for edge detection with a greater enhancement of images.

A new pattern of two dimensional cellular automata linear rules that are used for efficient edge detection of an image have proposed by Mohammed and Nayak [27]. They have observed four linear rules (namely $Rule_{29}$, $Rule_{113}$, $Rule_{263}$, and $Rule_{449}$) among $2^9$ total linear rules of a rectangular cellular automata in adiabatic or reflexive boundary condition that produces an optimal result. These four rules are directly applied to the images and produced edge detected output. Their results demonstrated that the proposed method gives better edge detection of an image than most of the traditional methods in terms of contrast enhancement which will be suitable for further analysis.

## 4.6 Noise reduction

Image noise is the unwanted information of an image (or the random disturbances in the images). Noise can occur during image capture, transmission or processing. Noises will reduce the quality of images and damage the expression of information for images. Image filtering is an important technique in the pre- processing of digital images [5]. This can effectively reduce the noise and make the image smooth. The corruption by impulse noise is a frequently encountered problem in image acquisition and transmission. Impulsive noise can be Salt and Pepper Noise (SPN) or Random Valued Impulsive Noise (RVIN). The pending problem that research in Random Valued Impulsive Noise (RVIN) filtering has been facing is the inability to distinguish noisy values that do not occur as extreme outliers in comparison with surrounding pixels.

There are many methods for reducing impulse noises such as the median filter which is a well-known and widely used conventional method. For noise removal, the median filter replaces the value of the center point by the median value of the neighboring pixels. However, the lack of differentiation between the noisy and non-noisy pixels of the image causes damage to the non-noisy pixels of the image too. That is these methods will destroy the natural texture and important information in the image like the edges. Most soft computing methods reduce noises effectively but lack of robustness, high costs of computation, being time-consuming, high complexity, and the need for evaluating and tuning the parameters are some problems of these methods [28].

CA has been successfully applied by different researchers in the area of noise filtering in the last couple of years. This section reviews some of the efficient research works on noise filtering based on CA.

Rosin in 2006 [13] used the sequential floating forward search method for feature selection to select good rule sets for a range of tasks, namely noise filtering (also applied to grayscale images using threshold decomposition), thinning, and convex hulls. He considered only the salt and pepper noise which is added to binary image for filtering task. The result of the proposed filters outperforms some traditional filters.

Selvapeter and Hordijk presented an image noise filter based on cellular automata (CA), which can remove impulse noise from a noise corrupted image [30]. The algorithm is applicable to both binary and gray level images, whereas in others only binary images were considered. They used a Moore neighborhood on binary images, and a Von Neumann neighborhood on both binary and gray scale images. Fixed value boundary conditions are applied, i.e., the update rule is only applied to non-boundary cells. An impulse noise corrupted image is taken as the initial CA lattice configuration. To remove this noise from the images, they used a majority CA update rule. This rule is stated as follows: if the center pixel (cell) gray level is 0 or 255 (i.e., black or white), then the gray level that is the majority in the local neighborhood replaces the center pixels value. If none of the gray levels in the local neighborhood is a majority, then there is a tie. This can be dealt with either deterministically or randomly. In the deterministic rule, the center pixel is replaced by the gray level which is in a fixed position in its local neighborhood (e.g., the pixel directly above it). Obviously this choice of the fixed position of the replacement pixel is arbitrary (it could also be the pixel directly below, or to the left, etc.). In the random majority rule, the center pixels value is replaced by the gray level of a randomly chosen pixel in its local neighborhood. In this case, the replacement pixel is chosen independently (at random) for each occurrence of a tie. The results using a cellular automaton based filter indeed show significant improvements over the performance of the various standard median filters.

Rosin's previous work mainly dealt with binary images, the current work operates on intensity images [14]. The increased number of cell states (i.e. pixel intensities) leads to a vast increase in the number of possible rules. Therefore, a reduced intensity representation is used, leading to a three state CA that is more practical. In addition, a modified sequential floating forward search mechanism is developed in order to speed up the selection of good rule sets in the CA training stage. Results are compared with our previous method based on threshold decomposition, and are found to be generally superior. The results demonstrate that the CA is capable of being trained to perform many different tasks, and that the quality of these results is in many cases comparable or better than established specialized algorithms.

Qadir et al. in 2012 [29] proposed an effective image noise filtering algorithm based on cellular automata, which is used to remove impulse noise from noise-corrupted images. In some traditional filtering algorithms, both corrupted as well as uncorrupted pixels are updated so thereby computationally inefficient and shows poor results. But, this method first identifies the corrupted pixels and replace them according to the proposed algorithm while the uncorrupted pixels remains unchanged. Two images of size 256 ×256 with varying percentage of salt and pepper noise were considered and results depicted that the proposed algorithm has significant improvements over the traditional filtering algorithms.

In 2012 Sadeghi et al. [28] used a hybrid method based on cellular automata (CA) and fuzzy logic called Fuzzy Cellular Automata (FCA) to eliminate impulse noises from noisy images in two steps. The type of impulse noise used here for experiment was random valued impulse noise. In the first step, based on statistical information, noisy pixels are detected by CA; then using this information, the noisy pixel will change by FCA. For noise detection and noise estimation, they used Von- Neumann neighborhood model. The proposed hybrid method is characterized as simple, robust and parallel which keeps the important details of the image effectively. The proposed approach has been performed on well-known gray scale test images and compared with other conventional and famous algorithms, is more effective.

### 4.7 Image Compression
Image compression deals with techniques for reducing the storage required to save an image, or the bandwidth required to transmit it. Paul et al. have proposed a new Cellular Automata based transform coding scheme for gray level and color still images [24]. This new scheme was found to be markedly superior to the currently available DCT based schemes both in terms of Compression Ratio and Reconstructed Image Fidelity. Cellular automata potential within image interpolation and image compression was investigated by Zhao et al. [32]. A specific class of CA, termed as Multiple Attractor Cellular Automata (MACA), has been evolved through Genetic Algorithm (GA) formulation to perform the task of image compression [31]. Extensive experimental results demonstrated better performance of the proposed scheme over popular classification algorithms in respect of memory overhead and retrieval time with comparable classification accuracy. Hardware architecture of the proposed classifier has been also reported.

## 5. EXPERIMENTAL RESULTS
Some of the algorithms of the above applications are implemented and the results are shown in this section. All the algorithms are implemented in MATLAB 2012a.

The following results as shown in Figure 6 illustrates the translation of an image of size 246×256 [9].

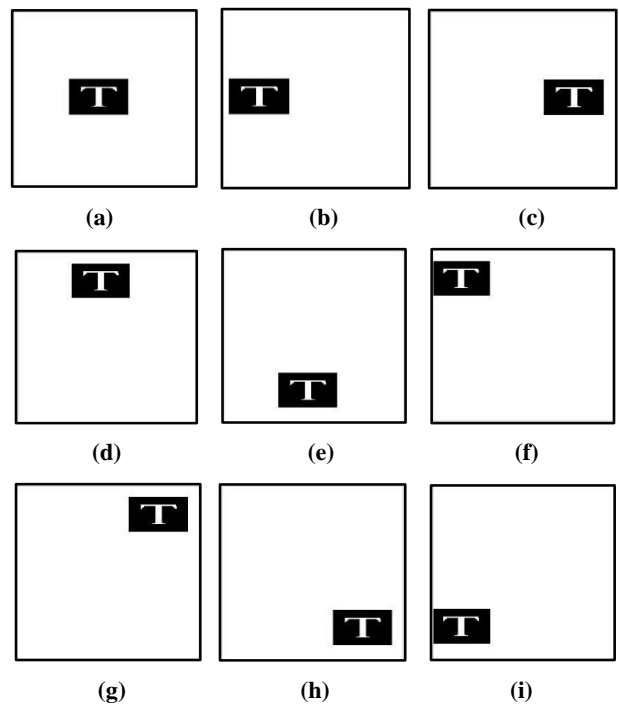

**Fig 6: Translation of an image of size 246×256 using 2D CA rules with 80 iterations (a) Original Image, (b) Rule$_2$ (Left), (c) Rule$_{32}$ (Right), (d) Rule$_8$ (Top), (e) Rule$_{128}$ (Bottom), (f) Rule$_4$ (Top- Left), (g) Rule$_{16}$ (Top- Right), (h) Rule$_{64}$ (Bottom-Right), (i) Rule$_{256}$ (Bottom-Left)**

Figure 7 and 8 demonstrates the action of zooming-in and zooming-out operations with the help of 2D cellular automata respectively [9].

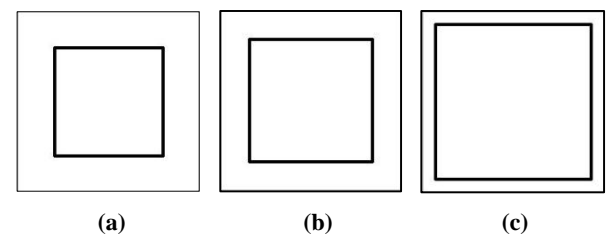

**Fig 7: Zooming-in operation using Cellular Automata (a) Original Image, (b) After 10 iterations, (c) After 20 iterations**

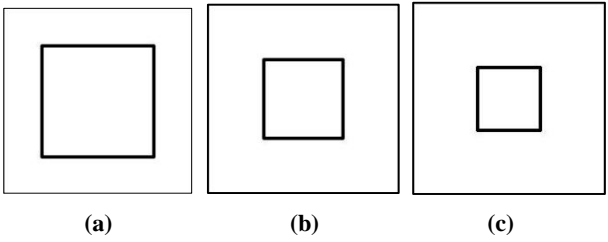

**Fig 8: Zooming-out operation using Cellular Automata (a) Original Image, (b) After 20 iterations, (c) After 30 iterations**

For edge detection we have implemented two algorithms employed by Nayak et al. [26, 27] and their results are shown in Figure 9. For this purpose, we take the experimental image as the X-ray image of size 302×270. Also we made a comparison of these algorithms with some traditional algorithms. The figure illustrates that the result produced by two algorithms are better in terms of contrast enhancement.

The results of the CA based filtering algorithm are demonstrated in Figure 10 [29]. In addition to that we have compared their results with some well-known algorithms and found that the CA based filter outperforms other filters. For performance comparison of different filters we used the measure PSNR (Peak Signal to Noise Ratio). PSNR (dB) is most easily defined via the mean squared error (MSE). PSNR is the measure of quality of an image. Hence, higher is the PSNR value and lower is the MSE value, better is the filter performance. Here, we consider a Lena image corrupted with Salt and Pepper Noise with noise density varies from 10%-50%. These noisy images are subjected to filtering by the proposed CA algorithm along with the some existing schemes such as Standard Median filter 3×3 (SMF3) [5], Standard Median filter 5×5 (SMF5) [5], Weighted Median Filter (WMF) with 5×5 window size [41], Center Weighted Median Filter (CWMF) with 5×5 window size and a center weight of 3 [42], Adaptive Center Weighted Median Filter (ACWMF) [43], Progressive Switching Median Filter (PSMF) [44], Decision based Algorithm for Impulse Noise (DBAIN) [45] and Noise Adaptive Fuzzy Switching Median Filter (NAFSMF) [46]. Table 4 shows the PSNR values for different filters and noise ratios for Lena image. From Table 4, we observed that the performance of the proposed filter in terms of PSNR (dB) is better than all of the schemes.

## 6. CONCLUSION

This paper presents a detailed survey in the domain of CA and its application in image processing. As CA is inherently parallel and having simple structure, it influences a large application in image processing. From the survey it has been found that most of the researchers used linear CA and very few of them are using hybrid and non-linear CA. Application of hybrid CA and non-linear CA is a challenging task for researchers in this field. The survey concludes that CA is found to be the parallel method to solve the any image processing problems. The extensive bibliography in support of the different developments of CA research provided with the paper should be of great help to CA researchers in the future.

## 7. ACKNOWLEDGMENTS

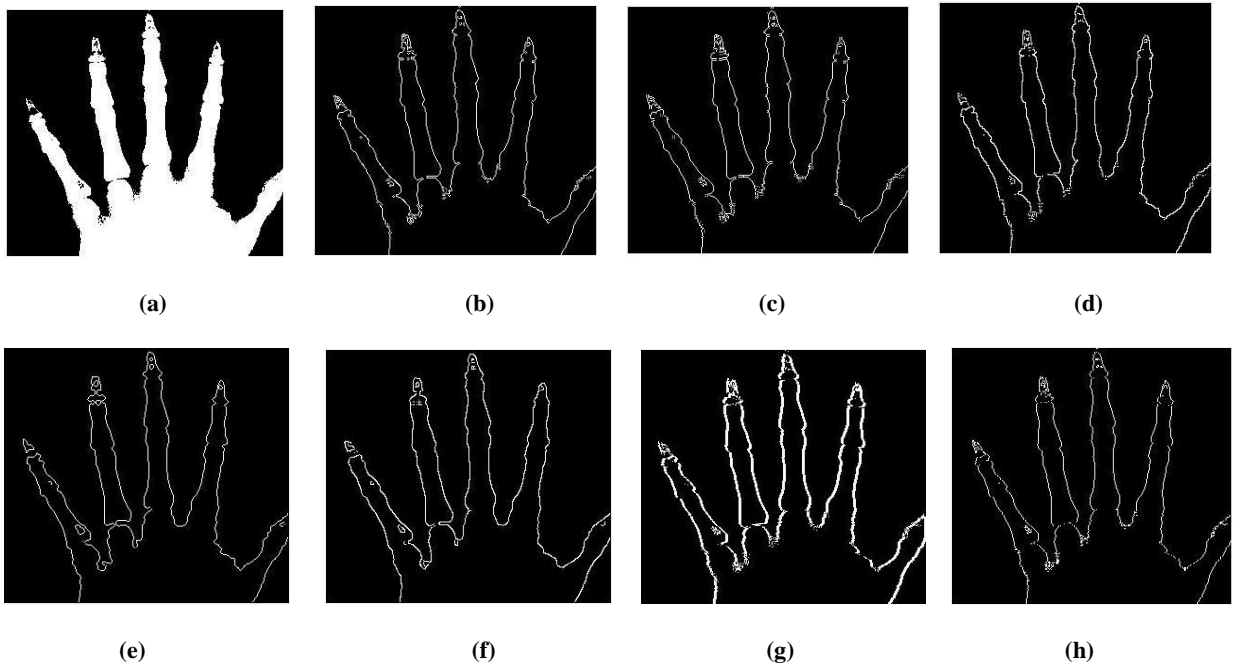

**Fig 9: Edge detection results of X-ray image using different methods (a) Original Image, (b) Sobel Method, (c) Prewitt Method, (d) Robert Method, (e) LoG Method, (f) Canny Method, (g) CA Method [26], and (h) CA Method [27]**

**Table 4. Comparative results in PSNR (dB) using different filters for Lena test image corrupted by Salt and Pepper (SPN) noise of varying strengths**

| Filters/Noise Ratio | 10% | 20% | 30% | 40% | 50% |
|---|---|---|---|---|---|
| SMF3 | 34.05 | 32.34 | 31.11 | 30.26 | 29.65 |
| `SMF5 | 32.98 | 31.68 | 30.64 | 29.84 | 29.18 |
| WMF | 33.55 | 32.07 | 31.01 | 30.11 | 29.49 |
| CWMF | 33.55 | 32.12 | 31.02 | 30.27 | 29.51 |
| ACWMF | 36.33 | 33.65 | 32.10 | 30.87 | 29.97 |
| PSMF | 36.12 | 33.52 | 32.01 | 30.91 | 30.02 |
| DBAIN | 37.01 | 33.99 | 32.30 | 31.00 | 30.04 |
| NAFSM | 37.10 | 34.12 | 32.37 | 31.06 | 30.09 |
| CA Filter | 37.21 | 34.17 | 32.49 | 31.57 | 30.64 |

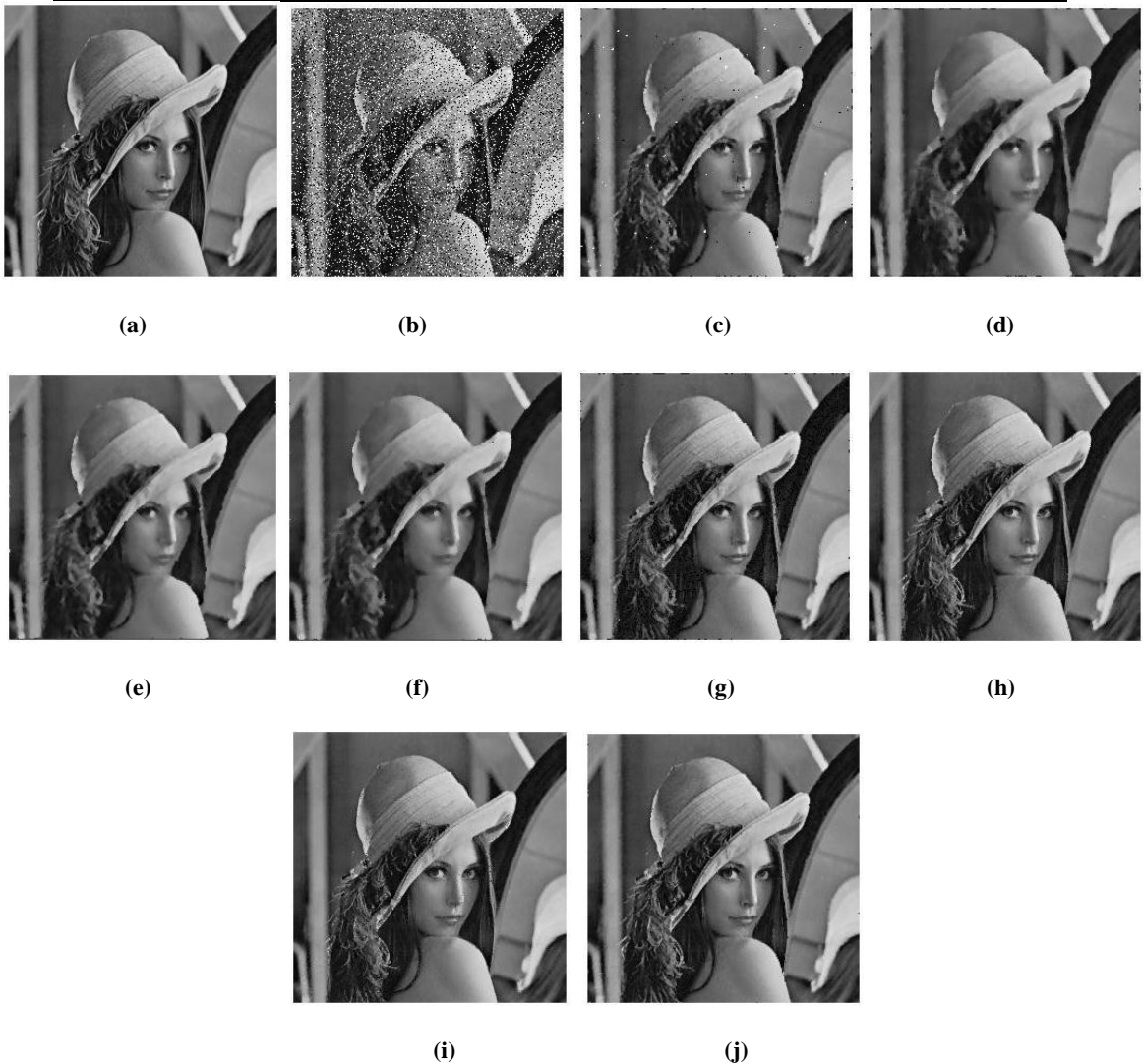

**(a)**       **(b)**       **(c)**       **(d)**

**(e)**       **(f)**       **(g)**       **(h)**

**(i)**       **(j)**

**Fig 10: Impulsive noise filtering of Lena image corrupted with 20% of SPN by different filters (a) Original Image, (b) 20% Noisy Image, (c) SMF3, (d) SMF5, (e) WMF, (f) CWMF, (g) PSMF, (h) DBAIN, (i) NAFSM, (j) CA Filter**